\documentclass[letterpaper, 10 pt, conference]{ieeeconf}  % Comment this line out if you need a4paper

\IEEEoverridecommandlockouts                              % This command is only needed if 
\usepackage{graphics} % for pdf, bitmapped graphics files
\usepackage{epsfig} % for postscript graphics files
\usepackage{mathptmx} % assumes new font selection scheme installed
\usepackage{times} % assumes new font selection scheme installed
\usepackage{amsmath} % assumes amsmath package installed
\usepackage{amssymb}  % assumes amsmath package installed
\usepackage{booktabs}
\usepackage{multirow}
\usepackage{algpseudocode}
\usepackage{float}
\usepackage{subfigure}
\usepackage[linesnumbered,ruled]{algorithm2e}
\usepackage{graphicx}
\usepackage{textcomp}
\usepackage{xcolor}
\usepackage{stackengine}

\usepackage{booktabs}
\usepackage{multirow}
\usepackage{algpseudocode}
\usepackage{float}
\usepackage{subfigure}
\usepackage[linesnumbered,ruled]{algorithm2e}
\usepackage{graphicx}
\usepackage{textcomp}
\usepackage{xcolor}
\usepackage{stackengine}
\usepackage{pgfplots}
\usepackage{pgfplotstable}
\usepackage{ifthen}
\pgfplotsset{compat=1.7}
\usepgfplotslibrary{groupplots}
\usepackage{verbatim}

\title{\LARGE \bf
Using Depth for Improving Referring Expression Comprehension \\in Real-World  Environments
}

\author{Fethiye Irmak Do\u{g}an$^{1}$ and Iolanda Leite$^{1}$% <-this % stops a space
\thanks{$^{1}$Fethiye Irmak Do\u{g}an, and Iolanda Leite are with the Division of Robotics, Perception and Learning from the School of Electrical Engineering and Computer Science at KTH Royal Institute of Technology, Stockholm, Sweden
        {\tt\small \{fidogan, iolanda\}@kth.se}}%
}

\begin{document}

\maketitle
\thispagestyle{empty}
\pagestyle{empty}

%%%%%%%%%%%%%%%%%%%%%%%%%%%%%%%%%%%%%%%%%%%%%%%%%%%%%%%%%%%%%%%%%%%%%%%%%%%%%%%%
\begin{abstract}
In a human-robot collaborative task where a robot helps its partner by finding described objects, the depth dimension plays a critical role in successful task completion. Existing studies have mostly focused on comprehending the object descriptions using RGB images. However, 3-dimensional space perception that includes depth information is fundamental in real-world environments. In this work, we propose a method to identify the described objects considering depth dimension data. Using depth features significantly improves performance in scenes where depth data is critical to disambiguate the objects and across our whole evaluation dataset that contains objects that can be specified with and without the depth dimension.

%In a human-robot collaborative task, when a robot helps its partner by finding the described objects, the depth dimension plays a critical role to successfully complete the task.
%When a robot follows the instructions of its human partner, it needs to perceive the world accurately for a successful collaboration. For instance, when its partner refers to an object in the environment using the relative positions, the robot needs to correctly identify the spatial relations between the objects to find out the described one. In real-world environments, depth dimension is critical while detecting these spatial relations and comprehending users' descriptions. 
%Existing studies has mostly focused on comprehending the object descriptions from RGB images. In this work, we propose a method to identify the described objects also considering the depth dimension. We present that using the depth features significantly improves the performance
%our method identifies the target object regions significantly better than the baseline 
%in the scenes where the depth is critical for the object descriptions and in our whole evaluation dataset containing objects that can be specified with and without the depth dimension.

\end{abstract}

%%%%%%%%%%%%%%%%%%%%%%%%%%%%%%%%%%%%%%%%%%%%%%%%%%%%%%%%%%%%%%%%%%%%%%%%%%%%%%%%
\section{Introduction}
\begin{figure}[t!]
\centering
\includegraphics[width=0.4\textwidth]{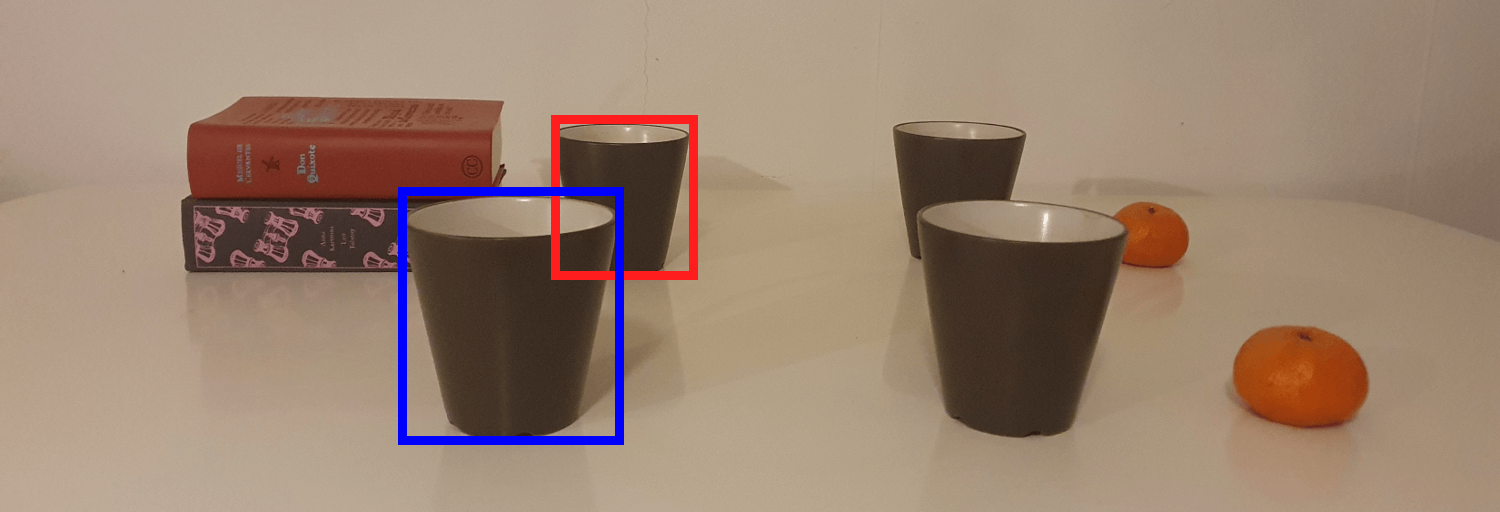}
\caption{An example illustrating the motivation behind using depth to improve referring expression comprehension. In this example, when the user's object description is `the mug next to the books', the robot can suggest the mug in the blue bounding box in RGB or the one in the red bounding box in RGB-D. Best viewed in color.}
\label{motivation}
\end{figure}

When a robot helps its human partner on a collaborative task, the depth dimension plays an important role for the robot to accurately comprehend the instructions of its partner.
For instance, consider a robot located in the environment of Figure \ref{motivation}, helping a user pick up a described object. In this scenario, if a user asks the robot to pick up `the mug next to the books', it can aim to take the incorrect mug (i.e., the one in the blue bounding box) using the RGB scene because this mug is the closest to the books in 2D. Alternatively, if it can obtain the RGB-D scene and use the depth dimension to solve the problem, the robot can aim to take the correct mug (i.e., the one in the red bounding box), which is the closest to the books in 3D. Therefore, the depth dimension is critical in this scenario to understand the user's object descriptions.

%When a robot works with a human partner on a collaborative task, it is critical to endow the robot with the ability to build the bridge between the visual input and verbal instructions of its partner. 
%To accurately build this bridge in real-world environments, the robot needs to exploit all the essential features that can be obtained from the visual input. For instance, if a user asks a robot to pick up a described object, the robot can aim to resolve the problem using a two-dimensional (RGB) or three-dimensional (RGB-D) representation of the visual input. 
%Further, in the user's object descriptions, the objects can be specified using their relative position to other objects (i.e., using the spatial relations between objects).  
%In this scenario, if the user specifies objects using their spatial relations (e.g., using their relative positions to other objects), the detection performance of these relations (e.g., `close by', `in front of', `next to') can be improved with the depth features~\cite{birmingham2018adding}. In this case, finding the described object can be impossible using the two-dimensional visual input without the depth features.

%In this real-world scenario, if the user's instruction contains spatial relations that can only be identified from the depth features (e.g., `close by', `in front of', `next to', `behind', etc.), finding the described object can be impossible without the depth dimension, which has been shown to improve the detection of spatial relations~\cite{birmingham2018adding}.

While describing objects, the expressions specifying them with their distinguishing features (such as their color, shape, or spatial relations) are called referring expressions. While comprehending these expressions, most techniques in computer vision and robotics studies have relied on flat RGB images without using the depth dimension~\cite{hatori2018interactively,yu2018mattnet,magassouba2019understanding,dogan2021}. However, depth information plays a critical role in real-world environments, and it was recently shown that depth features can facilitate the comprehension of referring expressions~\cite{Mauceri_2019_ICCV}. Consequently, there have been recent attempts to address this challenge 
%Comprehension of referring expressions from RGB images has been studied in many robotics applications~\cite{hatori2018interactively,yu2018mattnet,magassouba2019understanding,dogan2021}. Further, it has been recently shown that depth features can facilitate the comprehension of these expressions~\cite{birmingham2018adding}. Consequently, the problem has been also addressed 
using the three-dimensional feature space (i.e., 3D point clouds)~\cite{chen2020scanrefer,achlioptas2020referit_3d}. Although these studies have shown promising results, they have required candidate objects and selected the target object among the 3D object proposals. In contrast, our system addresses the challenge without this restriction by leveraging the explainability of image captioning -- see Section \ref{Heat} for further details. To our knowledge, our method is the first one to use explainability in RGB-D images to identify the described object regions in 3D environments.

In this paper, we extend our previous work~\cite{dogan2021} by providing the depth features in the input space and evaluating how the system performance improves with that addition. We first generate the RGB and depth activation heatmaps from the Grad-CAM explainability method~\cite{selvaraju2017grad}. Then, we obtain the combined activations showing the areas that are active in both of these heatmaps. 
Finally, we cluster the combined activations 
to generate suggested regions belonging to the described object.
%to find the active clusters, which are suggested as the regions belonging to the described object.
Our results show that depth features enhance the performance in the scenes where the object descriptions are dependent on the depth dimension and in the whole evaluation dataset.

\begin{figure*}[t!]
\centering
\includegraphics[width=0.955\textwidth]{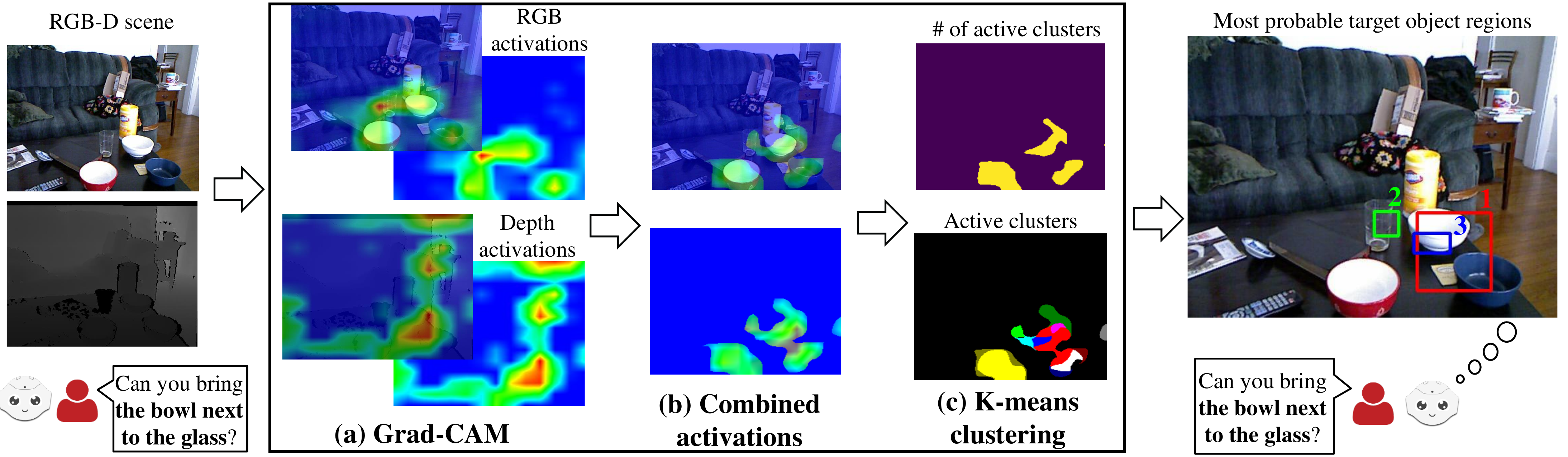}
\caption{For a given RGB-D scene and a referring expression (i.e., the bold part of the user expression), the overview of our suggested system to obtain the bounding boxes containing the target object regions.}
\label{fig:overview}
\end{figure*}

\subsection{Background}

Understanding users' object descriptions has long been a  consideration of various robotics studies. To build the bridge between the language and the two-dimensional visual input, recent studies~\cite{hatori2018interactively, magassouba2019understanding} have combined the features obtained from Convolutional Neural Networks (CNNs)~\cite{lecun1998gradient} and 
Long Short-Term Memories (LSTMs)~\cite{hochreiter1997long} or used these features on training Generative Adversarial Networks (GANs)~\cite{goodfellow2014generative}. In our recent work~\cite{dogan2021}, we addressed this problem using the Grad-CAM explainability method~\cite{selvaraju2017grad}. 

For comprehending referring expressions ~\cite{zender2009situated,10.1007/978-3-319-46493-0_48,shridhar2018interactive,shridhar2020ingress} and understanding natural language instructions~\cite{paul2016efficient,mees2021composing,venkatesh2020spatial}, spatial relations have been commonly exploited. For instance, Shridhar et al.~\cite{shridhar2018interactive,shridhar2020ingress} proposed an R-LSTM component in their system to predict the relational expressions (e.g., \textit{`a red can of soda'}) in addition to S-LSTM component predicting the self-referential ones (e.g., \textit{`a red can of soda'}). Further, Nagaraja et al.~\cite{10.1007/978-3-319-46493-0_48} provided CNN features to LSTMs to model spatial relationships between a region and its context regions.

While identifying the spatial relationships among objects, depth information has been shown to improve the task performance~\cite{birmingham2018adding}. Consequently, studies on referring expression comprehension have also focused on resolving this problem in three-dimensional feature space. For instance, 3D Point Clouds were used as an input to select the target objects among the detected object candidates~\cite{chen2020scanrefer} or segmented 3D instances~\cite{achlioptas2020referit_3d}. Further, Mauceri et al.~\cite{Mauceri_2019_ICCV} proposed an RGB-D dataset with referring expressions and evaluated this dataset with proof-of-concept experiments. In their experiments, they modified the referring expression generation model of Mao et al.~\cite{mao2016generation} to take the depth dimension as an input in addition to RGB features. They also used this generation method for comprehension by maximizing the probability of generating the input expression for candidate bounding boxes. Their findings showed pioneering results for our work: additional depth features enhanced the model's performance. However, their method assumed that the candidate bounding boxes were given or could be obtained by object box proposal systems, but our method does not require any candidate proposals thanks to leveraging explainability of image captioning activations.

Explainability methods can provide more interpretable results showing the reasoning behind the system predictions.
These methods are critical for building trust and reliance in AI systems~\cite{siau2018building,Bussone2015}. Because of their significant impacts, explainable systems have been focused on varied research communities~\cite{Abdul2018}. %From the perspective of HRI
In HRI studies, they have been discussed in association with the perceived intelligence of robots~\cite{tabrez2019improving} and the users' trust in them~\cite{setchi2020explainable,edmonds2019tale}. For instance, Tabrez and Hayes~\cite{tabrez2019improving} showed that the perceived intelligence of the robot was higher when the reasons for its behaviors were explained in a Sudoku variant game. Further, Edmonds et al.~\cite{edmonds2019tale} showed that trust in robots could be affected by the form of explanations (e.g., visual or textual). 

%Because of its significant impacts,  explainable systems has been focused by varied research fields.

%For instance, Setchi et al.~\cite{setchi2020explainable} presented some of the open challenges for explainable robotics from the perspective of HRI %and stated different levels of explanations might be needed for different levels of autonomy. They also discussed 
%and discussed the importance of trust (e.g., real time sensing of trust, using it as a control variable, etc.) for explainable robotics. Siau and Wang~\cite{siau2018building} stated that systems that can explain their behaviours are critical to build the trust in AI. Further, Edmonds showed that the trust on robots can also be affected by the form of explanations (e.g., visual or textual explanations)~\cite{edmonds2019tale}. 

%\cite{sridharan2019towards,tabrez2019improving,setchi2020explainable}

In addition to the aim of suggesting transparent system predictions, explainability has also been used for advancing the systems' functioning~\cite{dogan2021,Hendricks2018,Ross2017,Selvaraju2019a,Li2018}. For instance, Selvaraju et al.~\cite{Selvaraju2019a} aligned the visual explanations obtained from Grad-CAM with the human attention heatmaps to improve task accuracy in image captioning~\cite{lin2014microsoft} and visual question answering~\cite{antol2015vqa} tasks. Further, in our prior work~\cite{dogan2021}, Grad-CAM visual explanations were used during the inference of image captioning to find the described regions in RGB images -- see Section \ref{baseline}. This work extends our former method and employs the Grad-CAM activation heatmaps to identify the described objects in RGB-D images.

%\cite{Hendricks2018,Ross2017,Selvaraju2019a,Li2018}

%importance of explainability
%\cite{BarredoArrieta2020,Abdul2018,Bussone2015,Wieringa2020}

%%%%%%%%%%%%Explainability in HRI
%Existing research on HRI has evaluated the importance of explainable AI on different tasks \cite{sridharan2019towards} such as learning to open medicine bottles \cite{edmonds2019tale}, color-based collaborative Sudoku variant task \cite{tabrez2019improving}, delivering objects to people or places, and following recipes to bake biscuit \cite{setchi2020explainable}.
%Employing explainability for human-robot interaction can provide more interpretable results for humans and it is stated that it increases the trust in AI systems \cite{siau2018building}.

%%%%%%%%Works on CV to improve performance
%\cite{Hendricks2018,Ross2017,Selvaraju2019a,Li2018}

\subsection{Contributions}
Our contributions in this work can be summarized as follows:
\begin{itemize}
    \item We have extended our recent work~\cite{dogan2021} to take the depth dimension as an input, and we identify the target object regions from RGB-D images by leveraging explainability. To our knowledge, this is the first work using explainability while considering the depth of the objects to find the described object regions.
    \item We show that using the depth dimension improves the performance in scenes where the target objects are described with the spatial relations dependent on the depth features and in the whole evaluation dataset, which contains object features both dependent and independent of the depth dimension.
\end{itemize}
%%%%%%%%%%%%%%%%%%%%%%%%%%%%%%%%%%%%%%%%%%%%%%%%%%%%%%%%%%%%%%%%%%%%%%%%%%%%%%%%

\section{Finding the Described RGB-D Scene Regions}
\label{method}

To obtain the described RGB-D scene regions, we propose to extend our previous method that only takes RGB as input. 
%a method using explainability. 
Section \ref{baseline} describes our previously proposed method and highlights the differences between the two approaches. In this section, we explain our overall procedure to find the described regions in RGB-D scenes for a given expression. (See Figure \ref{fig:overview}  for an overview.) In the rest of the paper, the method using the depth features is referred to as the RGB-D method, and our previous method without the depth features is called the RGB method.
%We first detect the active parts of RGB and depth dimensions using the Grad-CAM explainability method~\cite{selvaraju2017grad}, and then we find the common areas that are active in both RGB and depth. After identifying these areas, we use K-means clustering to obtain active clusters. Finally, we suggest the regions from the active clusters as the regions belonging to the described object -- see Figure \ref{fig:overview}  for the overview of our system.

\subsection{Obtaining Heatmap Activations}
\label{Heat}
To obtain the active parts of scenes, we use the image captioning module of Grad-CAM~\cite{selvaraju2017grad}. For a provided caption and a scene, this module generates a heatmap activation that highlights the scene's areas specified in the caption and shows the parts contributing to the output predictions.

The image captioning module of Grad-CAM uses NeuralTalk2~\cite{karpathy2015deep} captioning model. During training, NeuralTalk2 learns a rich feature space (e.g., spatial relations, affordances, color, and shape of the objects) not limited to object categories. Using these rich features with Grad-CAM explainability enables our system to highlight the important areas without putting any restriction on object categories. Further, this approach removes the limitation of selecting the target object among given candidates in the previous studies.

\begin{figure}
\centerline{
\subfigure[RGB image]{
	\includegraphics[width=0.14\textwidth]{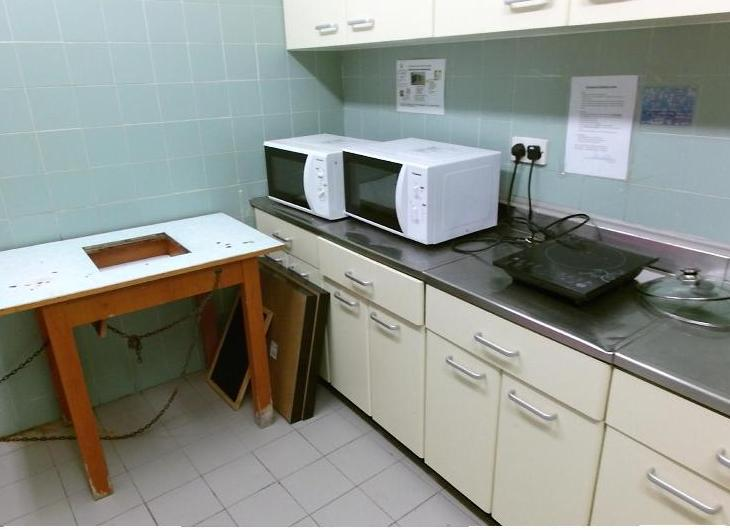}
	}
\subfigure[RGB activations]{
	\includegraphics[width=0.14\textwidth]{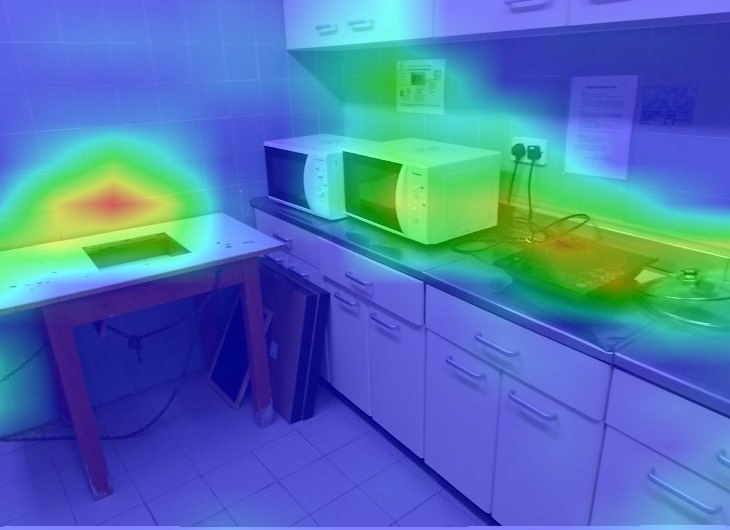}
	}
\subfigure[Depth activations]{
	\includegraphics[width=0.14\textwidth]{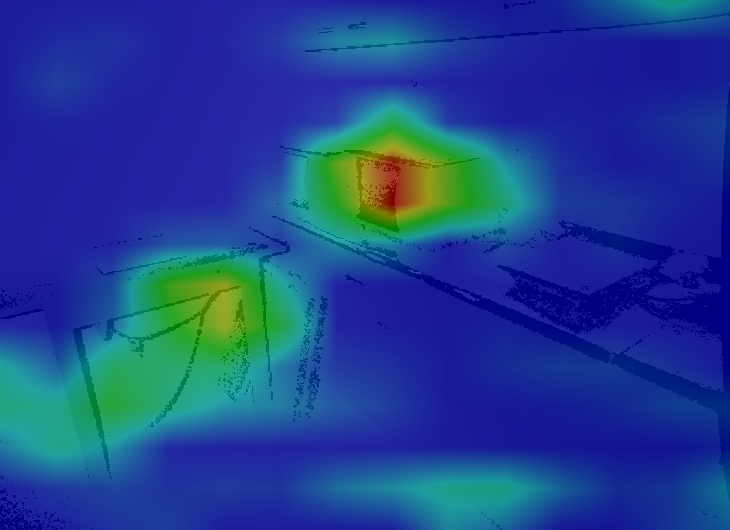}
	}
}
\caption{The heatmap activations of RGB image in (b) and depth activations in (c) when the expression is `the microwave closer to the table'.}
\label{fig:depth_grad}
\end{figure}

The NeuralTalk2 image captioning model was trained on RGB images, but thanks to its rich feature space, the Grad-CAM activations of the captioning model can also generate useful activations for the depth dimension of the scenes. For instance, in Figure \ref{fig:depth_grad}, heatmap activations of NeuralTalk2 in RGB image are not accurate enough to identify `the microwave closer to the table'. On the other hand, the heatmap of the depth image forces these activations towards the described area. Therefore, in this case, using the depth heatmap together with the RGB one can help to highlight the correct areas.

After observing the depth heatmap can help to identify the areas described by a user, as in Figure \ref{fig:depth_grad},  we provide an RGB-D scene to Grad-CAM through its RGB channels and depth dimension. Therefore, we obtain two different heatmaps, one from RGB denoted as $\mathcal{H}_{RGB}$ and another from the depth denoted as $\mathcal{H}_{depth}$. For instance, in Figure \ref{fig:overview}(a), the image in the back in the first row shows $\mathcal{H}_{RGB}$ and the image in the back in the second row visualizes the $\mathcal{H}_{depth}$.

In our heatmap representation, higher intensities in the red channel show higher activations, and higher values in the blue channels denote lower heatmap activations. We represent each pixel's normalized RGB channel intensities as $\{p^{RGB}_r,p^{RGB}_g,p^{RGB}_b\}$ and $\{p^{depth}_r,p^{depth}_g,p^{depth}_b\}$ for $\mathcal{H}_{RGB}$ and $\mathcal{H}_{depth}$ respectively.

\subsection{Combining Activations}
\label{combine}
After obtaining the activation heatmaps $\mathcal{H}_{RGB}$ and $\mathcal{H}_{depth}$, we find the intersecting area of the active parts in the heatmaps. First, we check the channel intensities of each pixel for both $\mathcal{H}_{RGB}$ and $\mathcal{H}_{depth}$. When red or green channel intensities are higher than a threshold $T_{rgb}$ (experimentally set as 0.39)  for both of the pixels in $\mathcal{H}_{RGB}$ and $\mathcal{H}_{depth}$, we assume that the corresponding pixel in their intersection heatmap $\mathcal{H}_{int}$ is also active. In that case, we take the mean of each channel in $\mathcal{H}_{RGB}$ and $\mathcal{H}_{depth}$ to set the corresponding pixel intensities $\{p^{int}_r,p^{int}_g,p^{int}_b\}$ in $\mathcal{H}_{int}$:
\begin{equation}
\label{eq1}
    p^{int}_r \leftarrow \frac{1}{2}({p^{RGB}_r + p^{depth}_r)},
\end{equation}
\begin{equation}
\label{eq2}
    p^{int}_g \leftarrow \frac{1}{2}({p^{RGB}_g + p^{depth}_g)},
\end{equation}
\begin{equation}
\label{eq3}
    p^{int}_r \leftarrow \frac{1}{2}({p^{RGB}_b + p^{depth}_b)}.
\end{equation}

If the red and green channels of a pixel in $\mathcal{H}_{RGB}$ or $\mathcal{H}_{depth}$ are lower than $T_{rgb}$, we set the corresponding pixel in $\mathcal{H}_{int}$ as inactive, i.e., we set $\{p^{int}_r,p^{int}_g,p^{int}_b\}$ as $\{0,0,1\}$ since the highest intensity in blue channel shows an inactive pixel. The second row of Figure \ref{fig:overview} (c) shows an example visualization of  $\mathcal{H}_{int}$.

\subsection{Clustering Heatmap}
\label{cluster}
After obtaining $\mathcal{H}_{int}$ showing the activation intersection of $\mathcal{H}_{RGB}$ and $\mathcal{H}_{depth}$, we cluster $\mathcal{H}_{int}$ to find the active regions in the RGB-D scene. To achieve this, we first obtain the number of clusters and then use this number for K-means clustering to identify the active clusters.

\subsubsection{Obtaining the number of clusters}

To obtain the number of clusters, we calculated the number of unconnected regions in $\mathcal{H}_{int}$. We first assign each pixel in $\mathcal{H}_{int}$ as either active or inactive. A pixel is assigned as active if $p^{int}_r$ or $p^{int}_g$ has a very high intensity value (i.e., higher than 0.9). Otherwise, it is labeled as inactive. Active pixels are labeled as one, and inactive pixel labels are set as zero. An example showing the visualization of the labeled pixels can be seen in the first row of Figure \ref{fig:overview}(c). After labeling each pixel as zero or one, we count the number of unconnected areas in the labeled image using pixels' 2D connectivity. While counting this number, denoted as $N$, we discard small unconnected areas (experimentally determined as smaller than 150 pixels) and also consider the background as an additional region. The computed number $N$ is given as the number of clusters to the K-means clustering.

\subsubsection{K-Means Clustering}
After finding the cluster count, we apply K-Means clustering to determine the active clusters. We first apply a Gaussian filter to $\mathcal{H}_{int}$ to smooth the pixel intensities of the heatmap. The filter's dimensions are set as 11, and the smoothed heatmap is represented as $\mathcal{H}_{S}$.

Then, we define a feature vector for each pixel in $\mathcal{H}_{S}$. After the smoothing, if a pixel is active (i.e., the red or blue channel has a value higher than 0.5), the feature vector of the pixel contains six different features:
\begin{equation}
\label{eq4}
    f_{p^{int}} \leftarrow \{p_x^{int}, p_y^{int},p_z^{int}, p_r^{int}, p_g^{int}, p_b^{int}\},
\end{equation}
{\noindent}where these features correspond to the pixel's coordinates in the x and y-axes, its corresponding depth value obtained from the input RGB-D scene, and its pixel intensities in red, blue, and green channels, respectively. All of these feature values are normalized in the zero to one range. Alternatively, if a pixel is not active after smoothing, the feature vector is set as $\{0,0,0,0,0,0\}$.

\begin{algorithm}[t!]
	\caption{The overall procedure to identify the described object regions.
		\label{alg:all_model}}
        \KwIn{
        an RGB-D scene and a referring expression
        }
        \KwOut{
         	the candidate bounding boxes showing the described object regions
        }
        Generate the heatmap activations $\mathcal{H}_{RGB}$ and $\mathcal{H}_{depth}$ using Grad-CAM\\
        Find the heatmap  $\mathcal{H}_{int}$ showing the common active areas of $\mathcal{H}_{RGB}$ and $\mathcal{H}_{depth}$ (Eq. \ref{eq1},\ref{eq2} and \ref{eq3})\\
        Count the number of unconnected areas ($N$) of active pixels in $\mathcal{H}_{int}$\\
        Obtain $\mathcal{H}_{S}$ by applying a Gaussian filter to $\mathcal{H}_{int}$\\
        Collect the feature vector of each pixel in $\mathcal{H}_{S}$ (Eq. \ref{eq4})\\
        Find the clusters by employing K-means clustering to the feature vectors with $N$ number of clusters\\
        Compute the activation of each cluster (Eq. \ref{eq5})\\
        Sort the clusters from the highest activation to the lowest activation\\
        Find the smallest bounding boxes covering the sorted clusters\\
        Provide the sorted bounding boxes as the candidate regions showing the target object regions\\
\end{algorithm}

Using the pixels' features and the calculated number of clusters $N$, we cluster the pixels of $\mathcal{H}_{S}$ with K-means clustering by minimizing the distance within clusters. After the convergence, the unconnected regions within the same clusters are considered separate clusters. Further, clusters with a small area (smaller than 150 pixels) are discarded from the obtained cluster list. Therefore, the final number of clusters can be different than the number provided to the K-means clustering algorithm. For instance, the number of clusters after K-means clustering shown in the second row of Figure \ref{fig:overview}(c) is more than the number of active clusters shown in the first row.

After the K-means clustering, we calculate the activation $a_{c_i}$ of each cluster $c_i$:
\begin{equation}
\begin{aligned}
a_{c_i} \leftarrow  \dfrac{1}{n_{c_i}}\sum_{\forall p^{int} \in c_i} (w_r\times p_r^{int} +w_g \times p_g^{int} ),\\ \ \textrm{for}\ c_i \in C \ \textrm{and}\ \ p^{int} \in \mathcal{H}_{int},
\label{eq5}
\end{aligned}
\end{equation}
{\noindent}where $C$ is the cluster list obtained from K-means clustering and $n_{c_i}$ is the number of pixels in $c_i$. Further, $w_r$ and $w_g$ are the weights showing the importance of the red and green channel intensities.  These values are experimentally determined as $0.7$ and $0.3$, respectively.

After obtaining $a_{c_i}$ for each cluster, we sort the clusters from the highest activation to the lowest. Then, we find the smallest bounding boxes covering these sorted clusters. Finally, we suggest the bounding boxes sorted with the same order of their corresponding clusters as the candidate bounding boxes containing the target object. Algorithm \ref{alg:all_model} summarizes the overall procedure of our system.

%%%%%%%%%%%%%%%%%%%%%%%%%%%%%%%%%%%%%%%%%%%%%%%%%%%%%%%%%%%%%%%%%%%%%%%%%%%%%%%%

\section{Experiments and Results}

\subsection{Finding the Described RGB Scene Regions}
\label{baseline}
To assess the impacts of depth features, we compared the RGB-D method (explained in Section \ref{method}) with our previous work (called the RGB method)~\cite{dogan2021}, %in, we selected our previous work as a baseline~\cite{dogan2021}. 
which uses Grad-CAM explainability to comprehend referring expressions on RGB scenes. The RGB method skips the steps explained in Section \ref{Heat} and \ref{combine}, and it obtains the heatmap activations providing a single RGB image and a referring expression to Grad-CAM. To find the active clusters and candidate bounding boxes from the heatmap, it follows the same procedure described in Section \ref{cluster}. However, in our previous formulation, the feature vector of a pixel shown in Eq. \ref{eq4} does not include the depth feature -- i.e., it only contains the pixel's x and y coordinates and the red, green, and blue channel intensities. Consequently, the K-means clustering is applied based on these five features. 

In the evaluation, the RGB method was compared with MAttNet~\cite{yu2018mattnet}, a state-of-the-art referring expression comprehension model. The results showed that compared to MAttNet, the RGB method performed better in the scenes with many distractors (i.e., the objects that are the same type as the target object) and uncommon objects that can't be identified with the exiting object detection methods (such as papaya and radish). Moreover, in our previous experiments, the results demonstrated that the regions proposed by the RGB method could be used for asking clarification questions to resolve the ambiguities.

\subsection{Data Collection}

To compare the RGB and RGB-D methods, we gathered a dataset with 70 scenes from SUN RGB-D~\cite{song2015sun}. This dataset contains various real-world scenes collected from different spatial contexts (e.g., living room, bedroom, bathroom, office, etc.). Moreover, for each scene, we selected a target object with at least one distractor  (i.e., the objects that are in the same object category as the target object). Further, for each target object, we collected an expression describing the target object in a natural and unambiguous manner. In the end, we obtained a dataset with 70 images and 70 expressions referring to the target objects.

Half of our dataset (35 images) was considered to be the easy category, and the remaining half was labeled as difficult. In the easy category, the target objects were described with features that were not tied to depth dimensions 
(e.g., the spatial relations such as `to the left', `to the right' or other object features such as the color or object type -- see Figure \ref{fig:image_results_easy} for some examples). In contrast, the difficult category images needed the depth dimension to disambiguate the target objects. Therefore, the expressions used to describe the target objects were dependent on their three-dimensional distances
(e.g., the expressions contained depth-dependent spatial relations such as `close by', `next to', `in front of', etc.) -- see Figure \ref{fig:image_results_diff} for some example images and expressions).

We collected such a dataset because we aim to assess the impacts of using depth features for dept dependent and independent environments. The easy and difficult category instances that we collected for this purpose enable us to manipulate the environment's depth dependence for a detailed comparison of the RGB and RGB-D methods. Moreover, the equal proportion of instances for each category ensures the fair evaluation of the methods' overall performance.

\subsection{Evaluation Procedure}

After obtaining the candidate bounding boxes from the RGB and RGB-D methods for each scene and expression, we reported the candidate bounding boxes that matched with the target objects. We evaluated the performances of both methods following the same procedure.

For each method's candidate bounding boxes, we computed a matching score between the bounding boxes and the target objects. To compute the matching score, we used the loss function $L_{DIoU}$ presented by Zheng and colleagues~\cite{zheng2020distance}. While computing the loss between two bounding boxes, $L_{DIoU}$ aims to minimize the distance between the bounding boxes' centers of mass and maximize their intersection area. %As in our previous work~\cite{dogan2021}, 
Using the loss $L_{DIoU}$, we obtained the matching score $M_{DIoU}$ with the following formulation:
\begin{equation}
    M_{DIoU} \leftarrow (1 - L_{DIoU}),
\end{equation}
{\noindent}where $M_{DIoU}$ varies from -1 to 1. A candidate bounding box is reported as matching the target object bounding box if $M_{DIoU}$ is higher than zero.

For each scene, we extracted the first three candidate bounding boxes suggested by each method. Then, we checked whether the first candidate matches with the target object. If the first candidate did not match, we checked the score for the second candidate bounding box. If none of the first three candidate bounding boxes matched with the target object, we reported these cases as none of the candidate bounding boxes matched with the target object.

\subsection{Results}

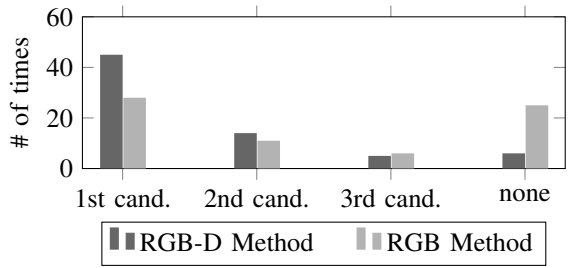
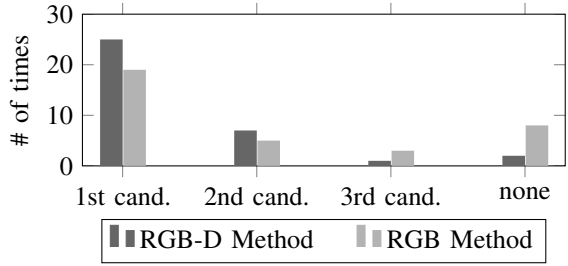
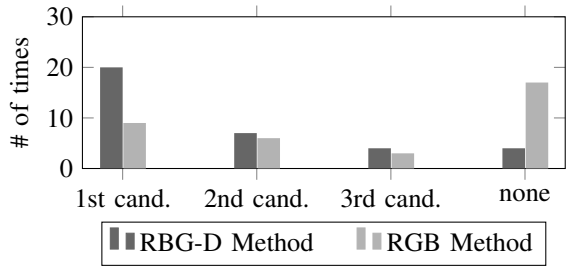
\begin{figure}[t!]
\subfigure[Whole dataset ($p=0.001$)]{
    \begin{tikzpicture}
\begin{axis}[
    legend style={at={(0.5,-0.35)},anchor=north, /tikz/every even column/.append style={column sep=0.5cm}},
    legend columns=3,
    height=3.6cm,
    width=8cm,
    ybar,
    ybar=0pt,
    enlarge x limits=0.1,
    ymin=0, ymax=60,%
    bar width=0.3cm,
    xtick=data,
    xticklabels={1st cand., 2nd cand., 3rd cand., none},
    ylabel style={align=center},
    ylabel= \# of times,
    %xlabel=correct candidate number,
    xticklabel style={align=center},
    cycle list name = exotic,
    every axis plot/.append style={fill,draw=none,no markers}
]
\addplot[color=black!60] table[x=condition,y=1]{plots/all.dat};
\addlegendentry{RGB-D Method}
\addplot[color=black!30, shift={(0.5,0)}] table[x=condition,y=2]{plots/all.dat};
\addlegendentry{RGB Method}
%\draw ([yshift=-1mm]axis cs:1,58) -- (axis cs:1,58) -- node[above, yshift=-1mm]{\small ***} (axis cs:4,58) -- ([yshift=-1mm]axis cs:4,58);
\end{axis}
\end{tikzpicture}
    \label{fig:all_results}}
\subfigure[Easy category ($p=0.12$)]{
    \begin{tikzpicture}
\begin{axis}[
    legend style={at={(0.5,-0.35)},anchor=north, /tikz/every even column/.append style={column sep=0.5cm}},
    legend columns=3,
    height=3.6cm,
    width=8cm,
    ybar,
    ybar=0pt,
    enlarge x limits=0.1,
    ymin=0, ymax=30,%
    bar width=0.3cm,
    xtick=data,
    xticklabels={1st cand., 2nd cand., 3rd cand., none},
    ylabel style={align=center},
    ylabel= \# of times,
    %xlabel=correct candidate number,
    xticklabel style={align=center},
    cycle list name = exotic,
    every axis plot/.append style={fill,draw=none,no markers}
]
\addplot[color=black!60] table[x=condition,y=1]{plots/easy.dat};
\addlegendentry{RGB-D Method}
\addplot[color=black!30, shift={(0.5,0)}] table[x=condition,y=2]{plots/easy.dat};
\addlegendentry{RGB Method}

\end{axis}
\end{tikzpicture}
    \label{fig:easy_results}}
\subfigure[Difficult category ($p=0.004$)]{
    \begin{tikzpicture}
\begin{axis}[
    legend style={at={(0.5,-0.35)},anchor=north, /tikz/every even column/.append style={column sep=0.5cm}},
    legend columns=3,
    height=3.6cm,
    width=8cm,
    ybar,
    ybar=0pt,
    enlarge x limits=0.1,
    ymin=0, ymax=30,%
    bar width=0.3cm,
    xtick=data,
    xticklabels={1st cand., 2nd cand., 3rd cand., none},
    ylabel style={align=center},
    ylabel= \# of times,
    %xlabel=correct candidate number,
    xticklabel style={align=center},
    cycle list name = exotic,
    every axis plot/.append style={fill,draw=none,no markers}
]
\addplot[color=black!60] table[x=condition,y=1]{plots/challenging.dat};
\addlegendentry{RBG-D Method}
\addplot[color=black!30, shift={(0.5,0)}] table[x=condition,y=2]{plots/challenging.dat};
\addlegendentry{RGB Method}
%\draw ([yshift=-1mm]axis cs:1,29) -- (axis cs:1,29) -- node[above, yshift=-1mm]{\small **} (axis cs:4,29) -- ([yshift=-1mm]axis cs:4,29);
\end{axis}
\end{tikzpicture}
    \label{fig:challenging_results}}
\caption{The number of times that the generated candidate bounding boxes matched with the target objects for the whole dataset, easy and difficult categories.}
\label{fig:results}
\end{figure}

%We evaluated the RGB and RGmethods with and without depth features
We compared the RGB-D method with the RGB method 
considering the number of times the target object matched with the candidate bounding boxes for different difficulty levels -- see Figure \ref{fig:results}. Further, we provided some qualitative examples showing the first candidate bounding boxes suggested by both methods for the easy (Figure \ref{fig:image_results_easy}) and difficult (Figure \ref{fig:image_results_diff}) categories.

\subsubsection{Whole dataset}

We first evaluated our results by considering the whole dataset (70 images). Figure \ref{fig:all_results} shows that the RGB-D method found the target object more often in its first and second candidates compared to the RGB method. Moreover, the cases where none of the first three candidates matched with the target object were rarer in the RGB-D method. Further analysis of these results with a Chi-Squared test showed that these differences were significant ($\chi^2\ (3,\ N = 140) = 16.06,\ p = .001 $; the mode is the first candidate for both methods, i.e., the candidate most often matched with the target object was the first candidate).

%We first evaluated our results by considering all images (70 images). Figure \ref{fig:all_results} shows that our method found the target object more often in its first or second candidate compared to the baseline. Moreover, the cases where none of the first three candidates matched with the target object was less in our method.

%To analyze the significance of these results, we ran a Chi-Square test: $\chi^2\ (3,\ N = 140) = 16.06,\ p = .001 $, the candidate most often matching with the target object was the first candidate for both methods -- i.e., the mode is the first candidate both for our method and the baseline. With the significance of all images' results, we observed that our method is significantly better than the baseline while suggesting the target object regions.

\subsubsection{Easy Category}

To assess the impacts of depth features, we also examined the results in the easy category (35 images), where the target object descriptions did not depend on depth. Figure \ref{fig:easy_results} shows that the RGB-D method's first and second candidates matched with the target object more often, and the RGB-D method failed less while suggesting the regions belonging to the target object. However, when we examined the results with Fisher's exact test (a Chi-Squared test could not be applied because some cells had a minimum expected value of fewer than five), we did not observe any significant differences between methods (Fisher's exact test value: $5.59$, $N = 70$, $p = 0.12$, the mode is the first candidate for both methods).

\subsubsection{Difficult category}
Finally, we evaluated the impacts of using the depth dimension for the difficult category (35 images), where the descriptions of the target objects' were tied to their depth features. The results shown in Figure \ref{fig:challenging_results} demonstrated that the regions identified by the RGB-D method in its first, second or third candidates matched with the target object more often compared to the RGB method. Further, the RGB-D method had fewer cases where none of its first three candidates matched the target object. To assess these results' significance, we ran another Fisher's exact test. The result of this analysis showed that the differences were significant (Fisher's exact test value: $12.67$, $N = 70$, $p = 0.004$; the mode is the first candidate for the RGB-D method and none of the first three candidates for the RGB method).

%To assess these results' significance, we ran another Fisher's exact test (Chi-Square test wasn't applicable due to the minimum expected value of each cell). The result of this analysis showed significant differences (Fisher's exact test value: $12.67$, $N = 70$, $p = 0.004$, the mode is the first candidate in our method and none of the first three candidates for the baseline). With these results, we observed that our method performs significantly better than the baseline for difficult images.

%%%%%%%%%%%%%%%%%%%%%%%%%%%%%%%%%%%%%%%%%%%%%%%%%%%%%%%%%%%%%%%%%%%%%%%%%%%%%%%%

\section{Discussion}

%Our quantitative results demonstrated that using the depth of the objects improved the system performance for the whole dataset and the difficult category, but it didn't affect the system performance for the easy ones. 
Our quantitative results demonstrated that for the easy category of images, using the depth of the objects did not affect the system performance. In this category, similar performances from the RGB and RGB-D methods were expected because the target object descriptions are not dependent on the depth dimension. However, the system performance was significantly improved for the whole dataset and the difficult category. Further, the improvement was even more distinct for the difficult instances. The performance advancements in the difficult category, which was collected to simulate depth-dependent environments,
%,where the depth-dependent features were used to disambiguate the target objects, 
show that considering depth is critical in real-world applications of referring expression comprehension.
In these applications, the objects are located in three-dimensional feature space, and finding the described object can be impossible without their depth features.
%, in which the objects are located in three-dimensional feature space. 
In such cases, when the robot is comprehending the user's expressions, the RGB-D method can be used for successful human-robot collaboration.

%our findings indicate that utilizing depth to comprehend users' expressions is fundamental for successful human-robot collaboration.

%Furthermore, the difference between the methods for the number of failures (i.e., none of the first three candidates matched with the target object) was more distinct in the difficult category. 
%The significant performance improvement for the whole dataset also supports this deduction. 
%In short, our findings show that considering the depth of the objects while comprehending users' expressions favors the task accuracy of human-robot collaboration.

\begin{figure}[t!]
\centerline{
\subfigure[`the red pillow on the left of the sofa']{
	\includegraphics[width=0.20\textwidth,height=0.13\textheight]{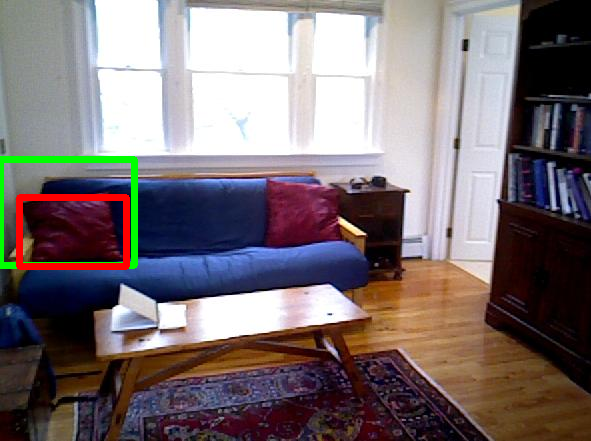}
	\includegraphics[width=0.20\textwidth,height=0.13\textheight]{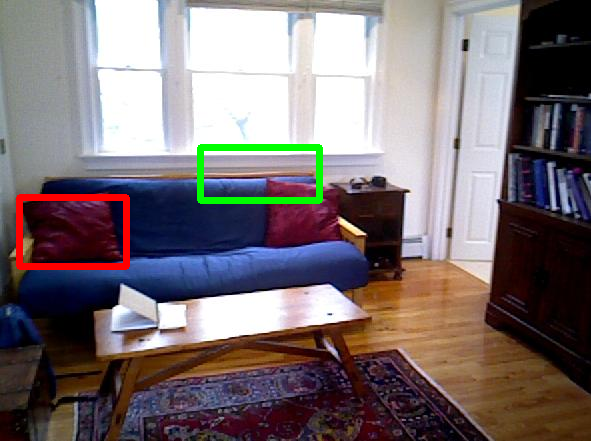}
	\label{e1}
	}
}

\centerline{
\subfigure[`the first towel from the left']{
	\includegraphics[width=0.20\textwidth,height=0.13\textheight]{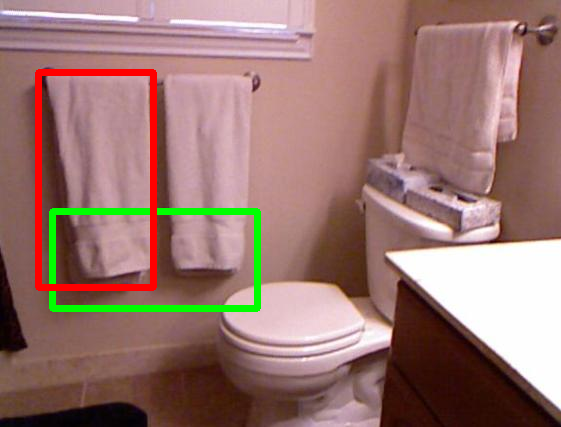}
	\includegraphics[width=0.20\textwidth,height=0.13\textheight]{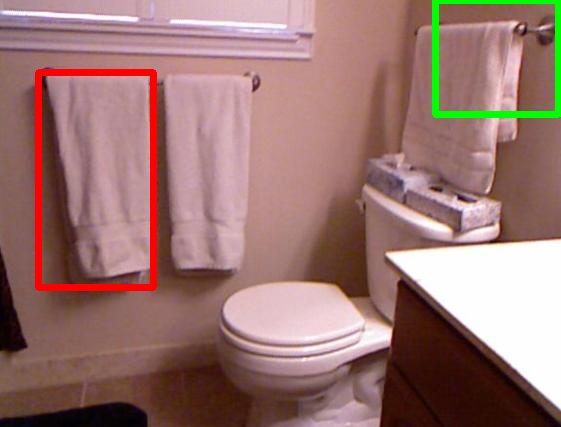}
	\label{e2}
	}
}

\centerline{
\subfigure[`the lamp to the right of the bed']{
	\includegraphics[width=0.20\textwidth,height=0.13\textheight]{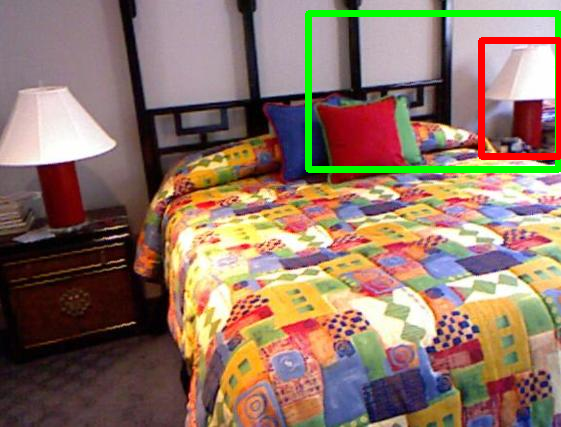}
	\includegraphics[width=0.20\textwidth,height=0.13\textheight]{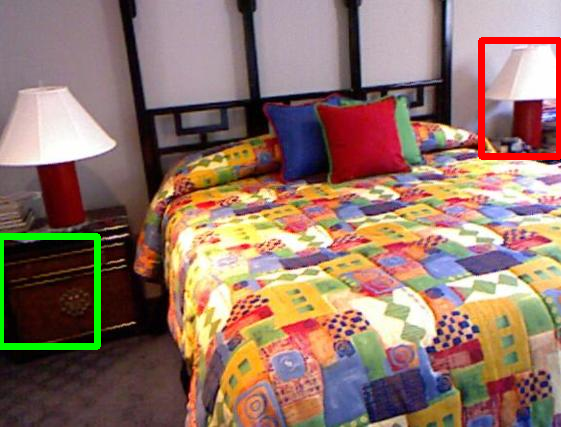}
	\label{e3}
	}
}

\caption{Examples from the easy category. The red bounding boxes show the target objects (ground truth), and the green boxes show the first candidates from the RGB-D method (left column) and the RGB method (right column) suggested for the given expressions. Best viewed in color.}
\label{fig:image_results_easy}
\end{figure}

\begin{figure}[t!]
\centerline{
\subfigure[`the chair in front of the fridge']{
	\includegraphics[width=0.20\textwidth,height=0.13\textheight]{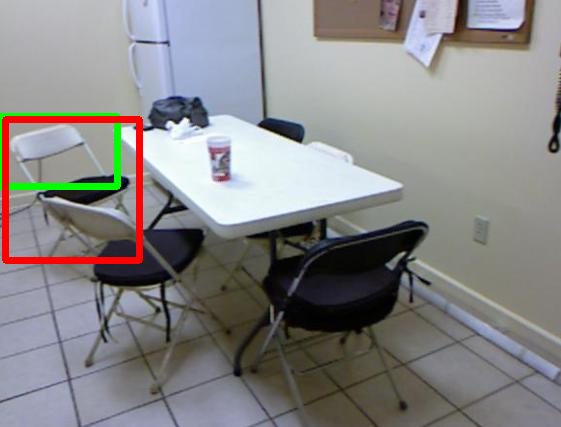}
	\includegraphics[width=0.20\textwidth,height=0.13\textheight]{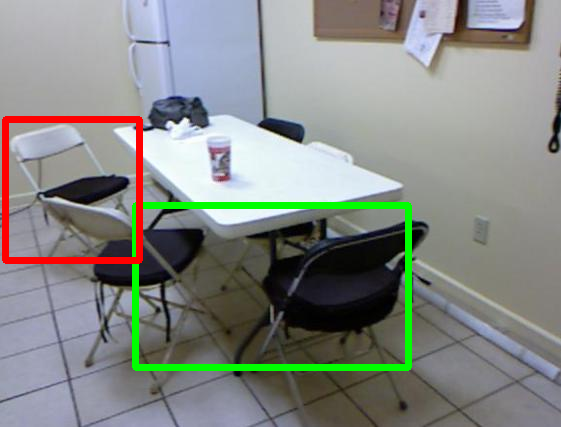}
	\label{d1}
	}
}

\centerline{
\subfigure[`the monitor next to the keyboard']{
	\includegraphics[width=0.20\textwidth,height=0.13\textheight]{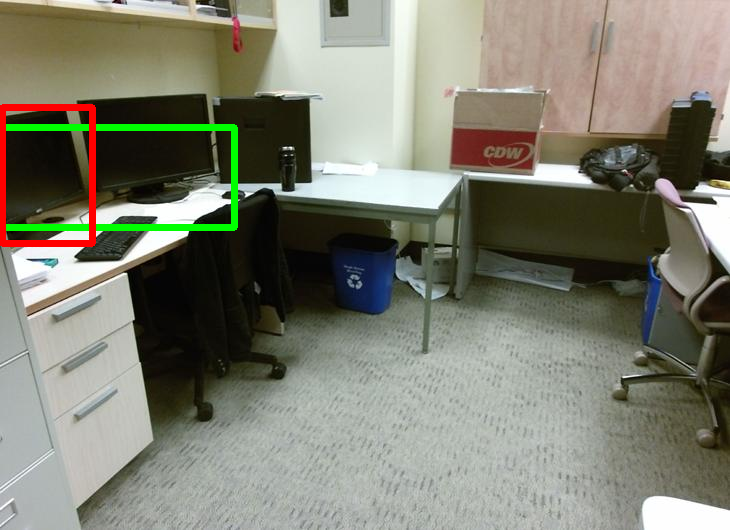}
	\includegraphics[width=0.20\textwidth,height=0.13\textheight]{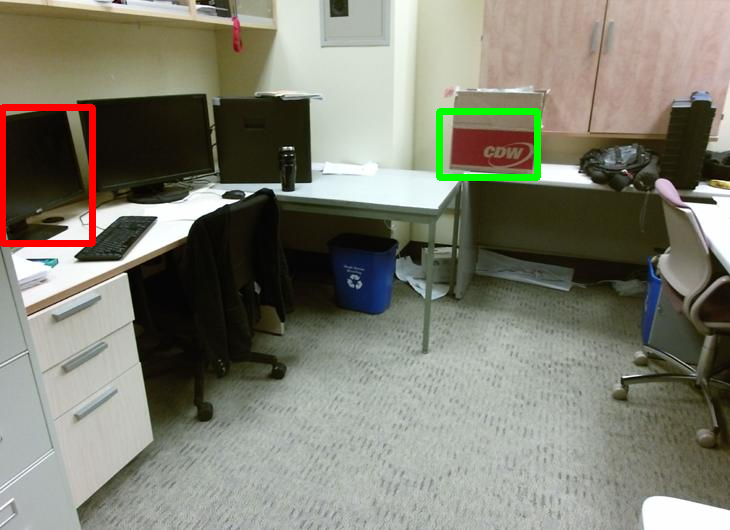}
	\label{d2}
	}
}
\centerline{
\subfigure[`the sofa in front of the window']{
	\includegraphics[width=0.20\textwidth,height=0.13\textheight]{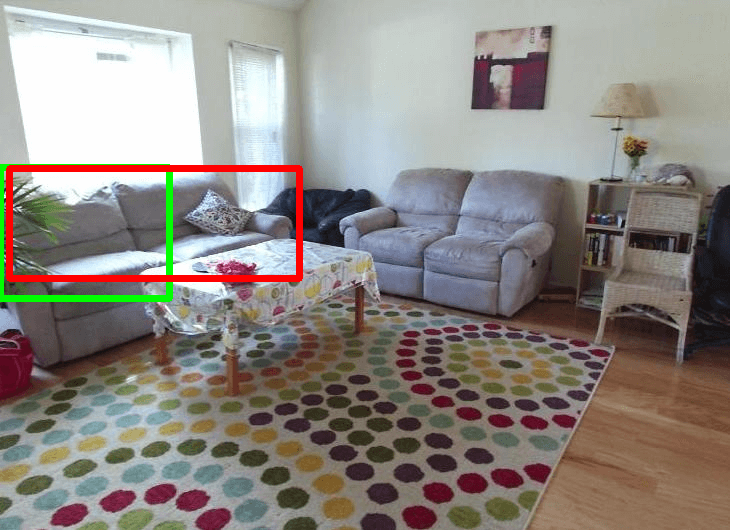}
	\includegraphics[width=0.20\textwidth,height=0.13\textheight]{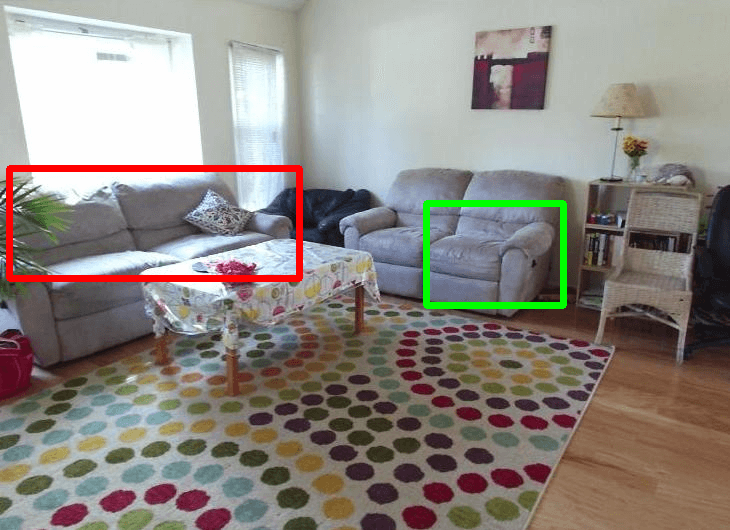}
	\label{d3}
	}
}

%\centerline{
%\subfigure[`the potted plant closest to the photocopy machine']{
%	\includegraphics[width=0.20\textwidth,height=0.13\textheight]{figures/bounds_depth_24.png}
%	\includegraphics[width=0.20\textwidth,height=0.13\textheight]{figures/bounds_24.png}
%	\label{d4}
%	}
%}

\caption{Examples from the difficult category. The red bounding boxes display the ground truth (target objects) for the given expressions, and the green boxes show the proposed first candidates from the RGB-D method (left column) and the RGB method (right column). Best viewed in color.}
\label{fig:image_results_diff}
\end{figure}

Our quantitative results also demonstrated that the RGB-D method could identify the target objects in its first candidate for the whole dataset and the difficult category more often than the RGB method could. Furthermore, the number of failures (i.e., none of the first three candidates matched with the target object) was significantly fewer for the RGB-D method in these cases. These findings imply that, in a real-world environment, the robot would find the described objects more often in its first selection without opting for its latter candidates, and it would make fewer mistakes if the depth dimension were provided in its input space. This suggests that using depth while comprehending users' expressions improves the task accuracy and efficiency of human-robot collaboration.%is crucial for the task accuracy and efficiency of human-robot collaboration. 

In our qualitative results from the easy category, we show the first candidate bounding boxes suggested by both methods in Figure \ref{fig:image_results_easy}. Even though we did not observe significant differences in our quantitative results for this category, Figure \ref{fig:image_results_easy} shows some of the examples in which
the RGB-D method (left column) suggested the regions matching the described objects better than the RGB method (right column). Although some bounding boxes from the RGB-D method do not exactly cover the target objects, the suggested regions are still sensible. For instance, the region suggested in Figure \ref{e3} partially contains the lamp and the bed when the expression is `the lamp to the right of the bed'. However, the region suggested by the RGB method is towards the incorrect lamp. Therefore, significant differences between methods for this category might be obtained with further analysis of the suggested regions by using different matching scores or asking users to evaluate these proposed regions.

In our qualitative results for the difficult category  (Figure \ref{fig:image_results_diff}), we show the first candidate bounding boxes obtained from the RGB-D (left column) and RGB methods (right column). We observe that the regions suggested by the RGB-D method fit better to the target object. In these examples, the lack of depth features misleads the RGB method to select the distractor objects. For example, in Figure \ref{d1}, when the expression is `the chair in front of the fridge', the RGB method highlighted the incorrect chairs, which can be considered in front of the fridge in 2D. However, the RGB-D method can handle these situations using the additional features obtained from the depth dimension. These examples demonstrate the significance of the depth features for accurate comprehension of referring expressions in real-world environments.

%%%%%%%%%%%%%%%%%%%%%%%%%%%%%%%%%%%%%%%%%%%%%%%%%%%%%%%%%%%%%%%%%%%%%%%%%%%%%%%%
\section{Conclusion and Future Work}
%while a robot works with a human partner on a collaborative task, utilizing depth is crucial for accurate comprehension of users' instructions. 

In this paper, we present a method to find the described object regions in RGB-D images. The method generates the activation heatmaps of RGB channels and the depth dimension using the explainability module. The combined activations, obtained from the common active parts of the heatmaps, are clustered to find the active clusters showing the target object. Our experiments demonstrate that using the depth dimension significantly improves the performance in the difficult category and the whole evaluation dataset, which includes all of the easy and difficult category instances.

Our work can be broadened in different directions. For instance, instead of obtaining RGB and depth activations separately, the Grad-CAM module can be used to take the three dimensions (i.e., an RGB-D scene) as an input. In this case, the challenge can be finding a pre-trained image captioning network that performs well in 3D scenes to visualize the RGB-D gradient activations.
If these activations can be obtained, our system can be applied to them to obtain the described object regions.
Further, our system can be deployed to a robot, and 3D point clouds can be provided in the input space instead of RGB-D images. In this situation, the performance of the robot can be evaluated with and without depth features, and the interaction can be examined for the user's trust and reliance on the system predictions, which are critical measures for explainable robotics.

%%%%%%%%%%%%%%%%%%%%%%%%%%%%%%%%%%%%%%%%%%%%%%%%%%%%%%%%%%%%%%%%%%%%%%%%%%%%%%%%

\section*{Acknowledgment}
This work was partially funded by a grant from the Swedish Research Council (reg. number 2017-05189) and by the Swedish Foundation for Strategic Research. We are grateful to Liz Carter for her valuable comments.

%%%%%%%%%%%%%%%%%%%%%%%%%%%%%%%%%%%%%%%%%%%%%%%%%%%%%%%%%%%%%%%%%%%%%%%%%%%%%%%%

\bibliographystyle{IEEEtran}
\bibliography{references}

%%%%%%%%%%%%%%%%%%%%%%%%%%%%%%%%%%%%%%%%%%%%%%%%%%%%%%%%%%%%%%%%%%%%%%%%%%%%%%%%

\end{document}